\pdfoutput=1

\documentclass[11pt]{article}

\usepackage{acl}

\usepackage{times}
\usepackage{latexsym}

\usepackage[T1]{fontenc}

\usepackage[utf8]{inputenc}

\usepackage{microtype}

\usepackage{inconsolata}

\usepackage{graphicx}

\usepackage{boxedminipage} 
\usepackage{xcolor} 
\usepackage{graphicx}
\usepackage{multirow}
\usepackage{multicol}

\usepackage{array}
\usepackage{arydshln}

\usepackage{booktabs}

\definecolor{gred}{RGB}{219,68,55}
\definecolor{gblue}{RGB}{66,133,244}
\definecolor{gyellow}{RGB}{244,180,0}
\definecolor{ggreen}{RGB}{15,157,88}
\definecolor{ggrey}{RGB}{115,115,115}

\usepackage{soul}   

\usepackage{caption}
\usepackage{subcaption}
\usepackage{tabularx}

\usepackage{amsmath}

\usepackage{adjustbox}

\usepackage{ragged2e}

\usepackage{amsfonts} 

\usepackage{CJKutf8} 

\usepackage{diagbox} 

\usepackage{algorithm}
\usepackage{algpseudocode}

\usepackage{bm}

\usepackage{enumitem}

\title{Not All Preference Pairs Are Created Equal:\\A Recipe for Annotation-Efficient Iterative Preference Learning~\thanks{The work described in this paper is substantially supported by a grant from the Research Grant Council of the Hong Kong Special Administrative Region, China (Project Code: 14200719). }}

\author{
 \textbf{Sen Yang\textsuperscript{1}},
 \textbf{Leyang Cui\textsuperscript{2}}\footnotemark[2],
 \textbf{Deng Cai\textsuperscript{2}},\\
 \textbf{Xinting Huang\textsuperscript{2}},
 \textbf{Shuming Shi\textsuperscript{2}},
 \textbf{Wai Lam\textsuperscript{1}}
\\
 \textsuperscript{1}The Chinese University of Hong Kong
 \hspace{10pt}
 \textsuperscript{2}Tencent AI Lab
\\
\texttt{\{senyang.stu,nealcly.nlp,thisisjcykcd\}@gmail.com}
\\
\texttt{timxthuang@tencent.com} \hspace{15pt}\texttt{wlam@se.cuhk.edu.hk}
}

\begin{document}
\maketitle
\renewcommand{\thefootnote}{\fnsymbol{footnote}}
\footnotetext[2]{Correspondence to Leyang Cui.}
\begin{abstract}
Iterative preference learning, though yielding superior performances, requires online annotated preference labels.
In this work, we study strategies to select worth-annotating response pairs for cost-efficient annotation while achieving competitive or even better performances compared with the random selection baseline for iterative preference learning.
Built on assumptions regarding uncertainty and distribution shifts, we propose a comparative view to rank the implicit reward margins as predicted by DPO to select the response pairs that yield more benefits.
Through extensive experiments, we show that annotating those response pairs with \textit{small} margins is generally better than \textit{large} or \textit{random}, under both single- and multi-iteration scenarios.
Besides, our empirical results suggest allocating more annotation budgets in the earlier iterations rather than later across multiple iterations.
\end{abstract}

\section{Introduction}
\label{sec:intro}

Large language models (LLMs)~\citep{touvron2023llama,openai2024gpt4} have shown remarkable capabilities to understand and generate human languages, supporting applications such as question answering, coding, and psychological counseling. One of the keys to such success is to align LLMs with human-desired behaviors through preference learning. This is accomplished by annotating preference datasets and employing preference learning methods such as proximal policy optimization (\citealp{schulman2017proximal}, PPO) and direct preference optimization (\citealp{rafailov2023direct}, DPO).

\begin{figure}
    \centering
    \includegraphics[width=0.98\linewidth]{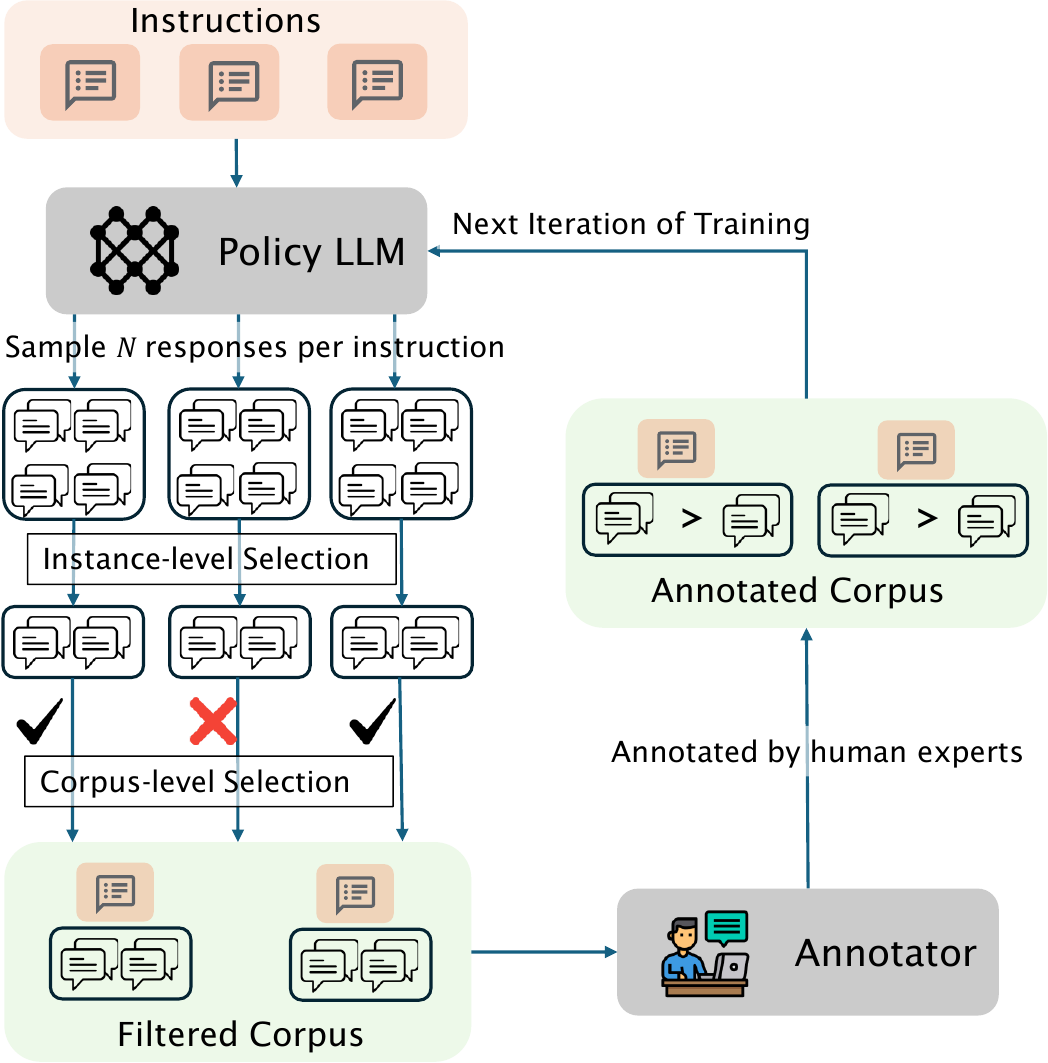}
    \caption{The workflow of online iterative preference learning, in which we apply two levels of selection before annotation.}
    \label{fig:framework}
\end{figure}

To continuously improve LLMs' capability,
recent work underlines the significance of iterative preference learning, which repetitively interleaves between training the model and collecting online preference annotations.
For example, LLaMA-2 and Claude series benefited from iterative RLHF training on human preference annotations that were collected in batches on a weekly basis~\citep{llama2,bai2022training}; while multiple papers also reported that iterative DPO brings clear performance gains~\citep{xu2024things,yuan2024selfrewarding,xiong2024iterative,rosset2024direct,wu2024selfplay}.

Despite their success, the process of collecting and annotating such online preference datasets is both time-consuming and costly.
These methods normally sample multiple responses per instruction on a large new collection of instructions and simply annotate all the responses.
The best and the worst responses are selected to formulate a pair per instruction to build a training corpus for the next iteration \cite{llama2,yuan2024selfrewarding,dong2024rlhf,wang2023helpsteer}.
This leads us to ask whether there exist alternative annotation strategies that are more cost-efficient.
Besides, existing methods normally allocate annotation budgets evenly across multiple iterations, but it remains unknown whether the model benefits from training on more instances in earlier or later iterations.
All in all, we are interested in a research question: \textit{how to make better use of limited annotation budgets} to aid online iterative preference learning.


In this paper, we address this question by conducting a systematic study on iterative DPO based on LLaMA-3-8B~\citep{llama3modelcard}.
We study the implicit reward margin as an informative indicator, which serves a key role in DPO and other direct preference learning methods ($\S$\ref{sec:single_iter}).
Such a choice is supported by the formulation of DPO, in which the reward margin roughly represents prediction uncertainty from the discriminative perspective or distribution shift from the generative perspective.
We consider two levels of granularity, namely the instance level and the corpus level, to rank reward margins from a comparative view.
Instance-level selection aims to find a worth-annotating pair of responses from $\frac{N(N-1)}{2}$ pairs if $N$ responses are sampled per instruction; while corpus-level selection considers filtering out those trios that do little help for alignment tuning from a large set of trios each consisting of an instruction and a pair of responses.


We conduct experiments in the single-iteration case, in which the policy LLM goes through one round of online training after being initially trained on an offline dataset.
We find that the \textit{smallest}-margin subset always works better than the \textit{largest}- and \textit{random}-subsets, on either the instance level or the corpus level.
Upon further checking the ranking accuracy and KL-divergences on each selected subset, we show that our assumptions regarding uncertainty and distribution shift are partly supported by our findings.
We then generalize the winning strategy, \textit{always-smallest}, to the multi-iteration case ($\S$\ref{sec:multi_iter}).
Experimental results demonstrate that the \textit{always-smallest} strategy yields continuous and significant improvements over multiple iterations; while the \textit{always-random} strategy sees little to no gains upon training on more iterations.
After that, we explore three strategies, namely \textit{increase}, \textit{constant}, and \textit{decrease}, to allocate annotation budgets to multiple iterations.
Empirical results suggest it is better to adopt \textit{decrease} and avoid \textit{increase}.


\section{Preliminaries}

\label{sec:setup}

In this section, we give a brief review of direct preference learning methods ($\S$\ref{sec:setup:dpo}) and iterative DPO ($\S$\ref{sec:setup:iterative_dpo}).
Then we present the preliminary setup of this work ($\S$\ref{sec:setup:preliminary}).

\subsection{Direct Preference Learning}
\label{sec:setup:dpo}

Recently, several works have bypassed the need to train a separate reward model, thus mitigating the instability issue of PPO training~\citep{dong2023raft,rafailov2023direct,zhao2023slichf}.
Among them, DPO gives a closed-form solution derived from the Bradley-Terry (BT) model~\citep{Bradley1952RankAO} to optimize a reward function from which the optimal policy is deterministically mapped:
\begin{align}
    \begin{split}
        & \mathcal{L}_{\mathrm{DPO}}(\pi_\theta, \pi_{\mathrm{ref}}) = - \mathbb{E}_{(x,y_w,y_l)\sim \mathcal{D}} \left[ \log \sigma ( \rho  ) \right]
    \end{split}
\end{align}
where
\begin{align}
    \rho & =  \beta \log \frac{\pi_{\theta}(y_w|x)}{\pi_{\mathrm{ref}} (y_w|x)} -  \beta \log \frac{\pi_{\theta}(y_l|x)}{\pi_{\mathrm{ref}} (y_l|x)}
\end{align}

Throughout this work, we use DPO as the default preference learning method due to its closed-form theoretical guarantee and stability during training.

\subsection{Online Iterative DPO}
\label{sec:setup:iterative_dpo}
Online iterative DPO is shown to be effective by multiple recent work~\citep{xu2024things,yuan2024selfrewarding,xiong2024iterative,rosset2024direct,wu2024selfplay,swamy2024minimaximalist,viethoangtranduong,ye2024online,guo2024direct,tajwar2024preference,calandriello2024human}.
Most of them repetitively interleave between training the model and collecting online preference annotations.

In this work, we assume a practical scenario where one round of supervised fine-tuning (SFT) and offline DPO has been implemented before the subsequent online iterations, given the availability of many open-sourced preference learning datasets~\citep{imdb,stienon2020learning,bai2022training,pmlr-v162-ethayarajh22a,nakano2022webgpt,h4stackexchange,cui2023ultrafeedback}.
Formally, we denote the SFT checkpoint as $\pi_{\mathrm{ref}}$ and the initial offline-tuned DPO checkpoint as $\pi_{\theta}^0$.
Our adopted iterative DPO framework repetitively applies the following Step-$i$:
    

\paragraph{Step-$\bm{i}$}
\begin{itemize}
    \item Given a set of $M$ instructions, $N$ responses are sampled from $\pi_{\theta}^{i-1}$ for each instruction. 
    $\pi_{\mathrm{ref}}$ and $\pi_{\theta}^{i-1}$ are used to predict the implicit reward\footnote{Such a definition is only valid under the case of reward margins so that the partition term is canceled. }, $\log \frac{\pi_{\theta}^{i-1}(y|x)}{\pi_{\mathrm{ref}} (y|x)}$.
    
    \item Some strategies are applied to select a subset of preference instances.
    The selected instances, each consisting of an instruction and two responses, are then annotated by an oracle preference annotator, e.g., human experts.

    \item The annotated instances are fed into $\pi_\theta^{i-1}$ and $\pi_\mathrm{ref}$ to train $\pi_\theta^i$.

\end{itemize}

Among all these sub-steps, our work focuses on how to select a proper subset of preference instances before annotation.
We present an illustration of the workflow in Figure~\ref{fig:framework}.

\section{Margin-based Selection within One Iteration}

\label{sec:single_iter}

We begin by analyzing the data selection strategies within a single iteration.
Existing methods~\citep{llama2,yuan2024selfrewarding,xiong2024iterative,rosset2024direct,wu2024selfplay} annotate all generated instances for the next iteration.
We question whether other strategies could achieve better performance with the same amount of annotation budgets.
We thus explore a simple yet intuitive metric for data selection, the reward margin between the chosen and the rejected responses.
The reward margin $\rho$ serves as the key component of the DPO loss function\footnote{$\rho$ is also the key role in many other direct preference learning methods, such as IPO (\citealp{azar2023general}, $\mathcal{L}_{\mathrm{IPO}}={(\rho - 1)}^2$) and SLiC (\citealp{zhao2023slichf}, $\mathcal{L}_{\mathrm{SLiC}}=\max\left\{ 0, 1-\rho \right\}$). }:
\begin{align}
    \begin{split}
        & \mathcal{L}_{\mathrm{DPO}}(\pi_\theta, \pi_{\mathrm{ref}}) = - \mathbb{E}_{(x,y_w,y_l)\sim \mathcal{D}} \left[ \log \sigma (  \rho  ) \right]
    \end{split}
\end{align}
where
\begin{align}
    \rho & = \underbrace{ \beta \log \frac{\pi_{\theta}(y_w|x)}{\pi_{\mathrm{ref}} (y_w|x)} -  \beta \log \frac{\pi_{\theta}(y_l|x)}{\pi_{\mathrm{ref}} (y_l|x)}}_\text{reward margin} \label{eq:margin} \\
    & =  \beta \underbrace{  \log \frac{\pi_{\theta}(y_w|x)}{\pi_{\theta} (y_l|x)}}_\text{policy log ratio} - \beta \underbrace{  \log \frac{\pi_{\mathrm{ref}}(y_w|x)}{\pi_{\mathrm{ref}} (y_l|x)}}_\text{reference log ratio} \label{eq:log_ratio}
\end{align}
There exist multiple interpretations for $\rho$.
The most straightforward one is to regard $\rho$ as the reward margin between $y_w$ and $y_l$ as predicted by a pairwise RM.
However, $\rho$ is much more intriguing than that because it directly models the output logits of $\pi_\theta$ and $\pi_{\mathrm{ref}}$ as implicit rewards, which represent the mixture of the two generative distributions.

Instead of staring at a single margin value that provides little insight for data selection, we consider ranking a set of reward margins.
We assume the highest- or lowest-ranking subset might be of use to find the worth-annotating instances.
We do not have a formal theory to support this assumption, but there are some intuitions from two points of view, i.e., discriminative and generative, associated with the mixed nature of DPO that optimizes a pairwise discriminative function that is built upon the output logits of generative LLMs.
We will show empirical evidence to support these intuitions in $\S$\ref{subsec:rq1:analysis}.

\paragraph{Uncertainty}
Upon regarding the policy model as a pairwise discriminative model, reward margins reflect the model's confidence in the prediction, so the most or least confident instances might be of extra use~\citep{10.5555/1619410.1619452,schroder-etal-2022-revisiting}.
Specifically, \citet{bai2022training} showed that the calibration curve of a preference model roughly matches the logistic function, $\mathrm{acc} = 1/(1+e^{-\rho})$, for models ranging from $10^8$ to $10^{10}$ parameters, demonstrating that reward margin is a good proxy of uncertainty.


\paragraph{Distribution Shift}
As shown in Eq~(\ref{eq:log_ratio}), reward margins represent the difference between the log ratios of the policy and the reference model; such a difference might correlate with the degree of generative distribution shift from the reference model to the policy. 
For example, if $y_w$ and $y_l$ lie in a similar distribution to that of the dataset where $\pi_\theta$ was trained on, there should be a clear gap between the two log ratios since $\pi_\theta$ has been trained to yield a higher generative probability for $y_w$ than $y_l$ while $\pi_{\mathrm{ref}}$ has not.
On the contrary, $y_w$ and $y_l$ may have not been effectively learned by $\pi_\theta$ if the two log ratios were almost canceled and the margin is small, in which case $\pi_\theta$ and $\pi_{\mathrm{ref}}$ show similar generative behaviors to distinguish $y_w$ from $y_l$ 

Given the above assumption that it might be useful to rank reward margins, the question then becomes how to define a good set of preference pairs on which we can get an informative ranking.
In the next section, we will discuss two levels of granularity to define potentially useful sets for ranking.



\begin{figure*}[th!]
\captionsetup[subfigure]{justification=Centering}
\centering
\begin{subfigure}[t]{0.44\textwidth}
    \includegraphics[width=\linewidth]{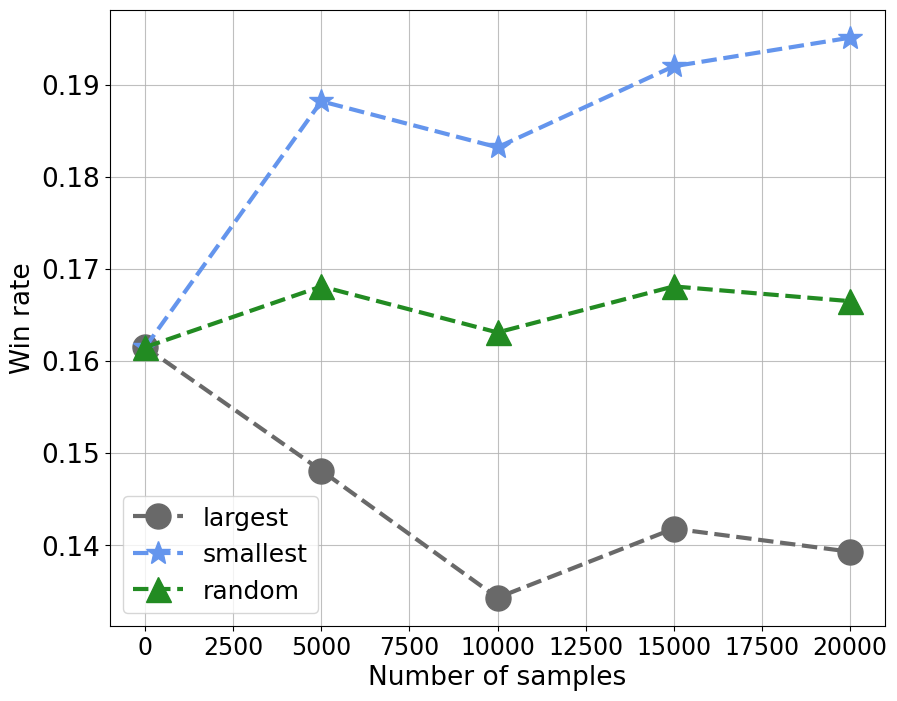}
    \caption{Evaluated the gold RM. } \label{subfig:exp:rq1:instance-pairrm}
\end{subfigure} \hspace{\fill} 
\begin{subfigure}[t]{0.44\textwidth}
    \includegraphics[width=\linewidth]{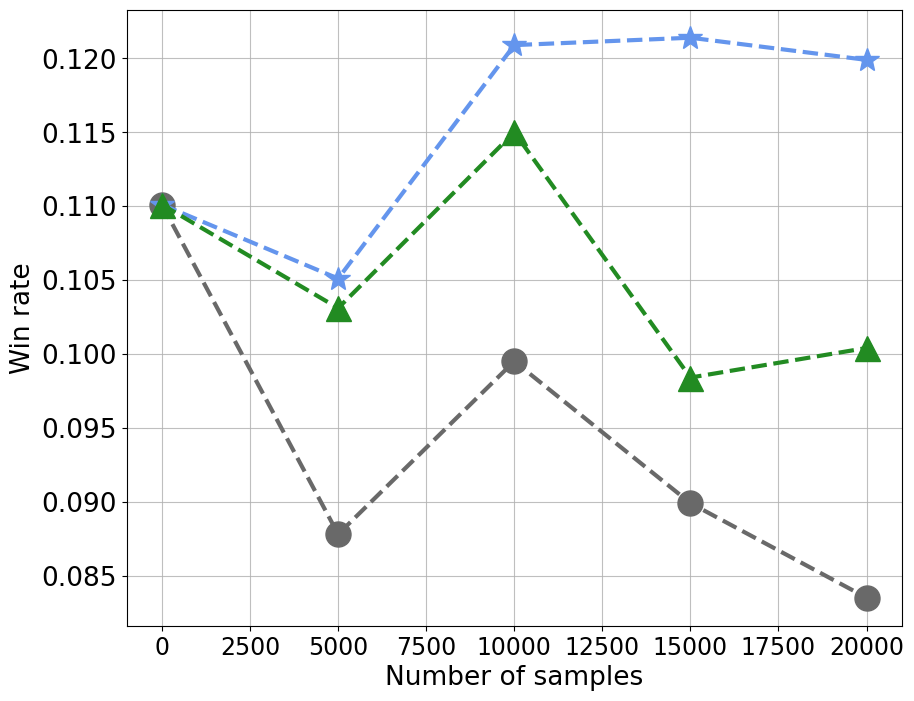}
    \caption{Evaluated by GPT-4. } \label{subfig:exp:rq1:instance-llama}
\end{subfigure}
\caption{
        Results on AlpacaEval-2.0 with different \ul{instance}-level strategies and different training set sizes.
        }
\label{fig:exp:rq1:instance}
\end{figure*}

\begin{figure*}[th!]
\captionsetup[subfigure]{justification=Centering}
\centering
\begin{subfigure}[t]{0.42\textwidth}
    \includegraphics[width=\linewidth]{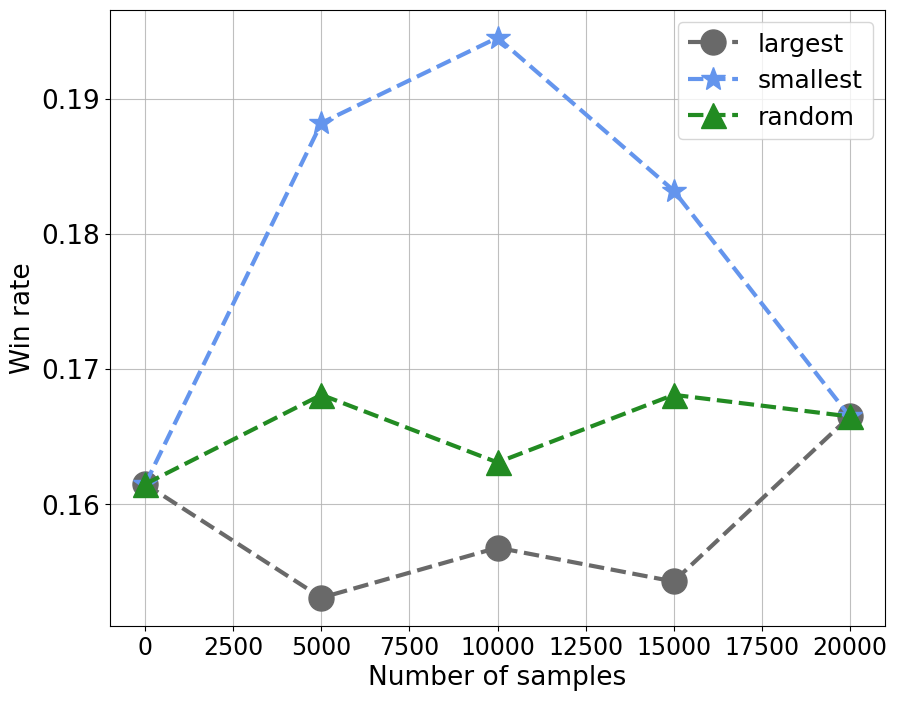}
    \caption{Evaluated by the gold RM. } \label{subfig:exp:rq1:corpus-pairrm}
\end{subfigure} \hspace{\fill} 
\begin{subfigure}[t]{0.42\textwidth}
    \includegraphics[width=\linewidth]{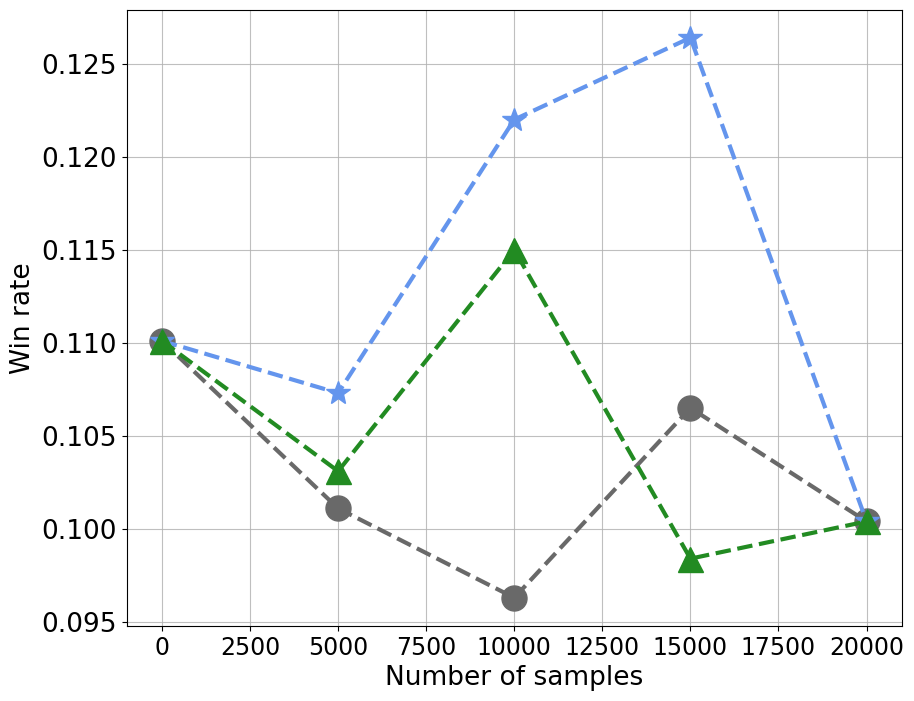}
    \caption{Evaluated by GPT-4. } \label{subfig:exp:rq1:corpus-llama}
\end{subfigure}
\caption{
        Results on AlpacaEval-2.0 with different \ul{corpus}-level strategies and different training set sizes.
        }
\label{fig:exp:rq1:corpus}
\end{figure*}

\subsection{Strategy Variants}

\label{sec:rq1_strategy}

We explore two levels of granularity to rank reward margins to select worth-annotating instances.
The instance-level selection ranks the margins of responses sampled from the same instruction; while the corpus-level selection applies to the set of response pairs of the entire corpus.


\paragraph{Instance-level}
Existing online preference learning methods that rely on human annotations normally send multiple responses for annotating per instruction and select the best and the worst to compose a pair.
However, it would be too costly to adopt this brute force strategy that asks a human expert to read and rank all $N$ responses.
We thus investigate the margins between any two responses to select a worth-annotating pair among $\frac{N(N-1)}{2}$ pairs, so that the cost remains the same as in $N=2$ while the diversity of responses is promoted.

\paragraph{Corpus-level}
Given a set of instances, each consisting of an instruction and a pair of responses, it is intuitive to discriminate between the instances that are beneficial from those less beneficial or even counterproductive.
We thus explore to rank the reward margins between pairs of responses over the entire corpus, seeking an informative partition.

The two levels of selection can be applied consecutively: One may first select the smallest-margin response pair for each instruction, then gather all such instances together to form a corpus, and finally select the largest-margin subset on the corpus level.
We do not consider all the combinations of the two levels due to prohibitive computational costs.
Instead, we assume one to be \textit{random selection} when experimenting with the other.

\paragraph{Margin Normalization}
The length bias issue has been shown to prevail among RLHF methods, including DPO~\citep{park2024disentangling,meng2024simpo}.
In our experiments, we also find the margin-based rankings of response pairs on the corpus level vary significantly between the length-normalized and un-normalized versions.
We thus consider length normalization to mitigate possible length bias during ranking.
Specifically, in addition to the un-normalized experiments, we also conduct experiments with the following normalized margin:
\begin{align}
    \begin{split}
        \hat{\rho} &= \frac{1}{|y_w|} \log \frac{\pi_{\theta}(y_w|x)}{\pi_{\mathrm{ref}} (y_w|x)} - \frac{1}{|y_l|} \log \frac{\pi_{\theta}(y_l|x)}{\pi_{\mathrm{ref}} (y_l|x)}
    \end{split}
\end{align}

\subsection{Experimental Setup}
\label{sec:setup:preliminary}

\subsubsection{Synthetic Oracle}
We aim to analyze annotation efficiencies in iterative preference learning, in which online preference annotations are collected in batches.
However, obtaining the true ``gold standard'' preference labels from human annotators can be costly, and may be inconsistent if the group of annotators varies across batches.
Inspired by \citet{Gao2022ScalingLF}, we instead employ a synthetic setup where the ground truth is determined by the outputs of a reliable reward model, which we term as \textit{gold RM}.
This gold RM is regarded as the alternative of human experts across all our experiments in terms of both annotation and evaluation, despite it is not the real ground truth.

\subsubsection{Training}

We adopt \textsc{LLaMA}-3-8b-base~\citep{llama3modelcard} to initialize the policy LLM and PairRM~\citep{llm-blender-2023} as the gold RM.
We sample 10,000 instances from UltraFeedback~\citep{cui2023ultrafeedback}, which are then used to train $\pi_{\mathrm{ref}}$ and $\pi_{\theta}^0$.
The instructions from the remaining UltraFeedback are kept for subsequent iterations.
We use $\pi_{\mathrm{ref}}$ as the reference model for DPO training across all iterations.
The training hyper-parameters are listed in Appendix \ref{sec:appendix:implmentation}.


\subsubsection{Evaluation}

We evaluate the policy LLMs on AlpacaEval-2.0~\citep{alpaca_eval} and use the outputs generated by GPT-4 as the reference to compare against.
We aim to get findings from the judgments predicted by the gold RM.
Such findings are approximations for the industrial scenario where human experts take the place of our gold RM.
One concern with this synthetic setup is the reward hacking issue, in which the policy overfits the preferences of the gold RM since the gold RM remains fixed after all~\citep{Gao2022ScalingLF,rafailov2024scaling}.
We investigate whether this issue exists in our experiments by additionally evaluating with the standard AlpacaEval-2.0 protocol, i.e., GPT-4\footnote{GPT-4-1106-preview} as the evaluator.
A model is deemed as ``hacked'' if it shows improvements when evaluated by the gold RM but degrades when evaluated by GPT-4.
It should be noted that the judgments predicted by the gold RM are still considered accurate proxies under the synthetic setup upon being hacked, though they would diverge from real human preferences.

Overall, we consider two criteria: (1) How well does the model align with the gold RM which is regarded as the ``ground truth'' in our experiments? (2) How does the model perform under general evaluation?



\subsection{Empirical Workflow}
Our workflow starts from $\pi_{\theta}^0$ and a set of 20,000 instructions sampled from UltraFeedback~\citep{cui2023ultrafeedback}:

\paragraph{Step 1}
$N=8$ responses are sampled from $\pi_{\theta}^0$ for each instruction. 
$\pi_{\mathrm{ref}}$ and $\pi_{\theta}^0$ are used to predict the implicit reward (without the partition term), $\log \frac{\pi_{\theta}^0(y|x)}{\pi_{\mathrm{ref}} (y|x)}$, and the normalized version, $\frac{1}{|y|}\log \frac{\pi_{\theta}^0(y|x)}{\pi_{\mathrm{ref}} (y|x)}$.

\paragraph{Step 2}
Margin-based strategies are adopted to select a subset of preference pairs.
The selected pairs are then annotated using the gold RM.
\begin{itemize}
    \item \textbf{Instance-level}: For each prompt, the preference pairs with the \{\textit{largest} \& \textit{smallest}\} margin  are selected among all the response pairs.
    A \textit{random} baseline is also included for comparison, in which the selected pair is simply the first two responses.
    All such prompt-chosen-rejected trios are collected to formulate a dataset.

    \item \textbf{Corpus-level}: Given a set of instances where each instance consists of a prompt and two responses, those pairs with the \{\textit{largest} \& \textit{smallest}\} margins are selected.
    Similarly, a \textit{random} baseline is included.

    \item \textbf{Margin Normalization}: Length normalization is enabled and disabled, respectively, for all the variants adopted.
\end{itemize}

\paragraph{Step 3}
The collected subset is fed into $\pi_\theta^{0}$ and $\pi_{\mathrm{ref}}$ to obtain the online trained policy, which then gets evaluated.

\begin{table}[t]
    \centering
    \begin{adjustbox}{max width=0.9\columnwidth}
    \begin{tabular}{l|cc}
        \hline
         \diagbox{\textbf{Ranking}}{\textbf{Selected}} & {Largest} & {Smallest} \\
         \hline
        {Instance-level} & 50.75$_{ \pm 0.51 }$ & 47.83$_{ \pm 2.10 }$ \\
        {Corpus-level} & 49.51$_{ \pm 1.15 }$ & 49.08$_{ \pm 2.26 }$ \\
        \hline
    \end{tabular}
    \end{adjustbox}
    \caption{Win rates of \ul{length-normalization} models against the \ul{un-normalized} counterparts with several variants, as evaluated by the gold RM. The win rates are averaged over multiple runs with different numbers of instances. }
    \label{tab:exp:rq1:avg_logp}
\end{table}

\begin{figure*}[th!]
\captionsetup[subfigure]{justification=Centering}
\centering
\begin{subfigure}[t]{0.26\textwidth}
    \includegraphics[width=\linewidth]{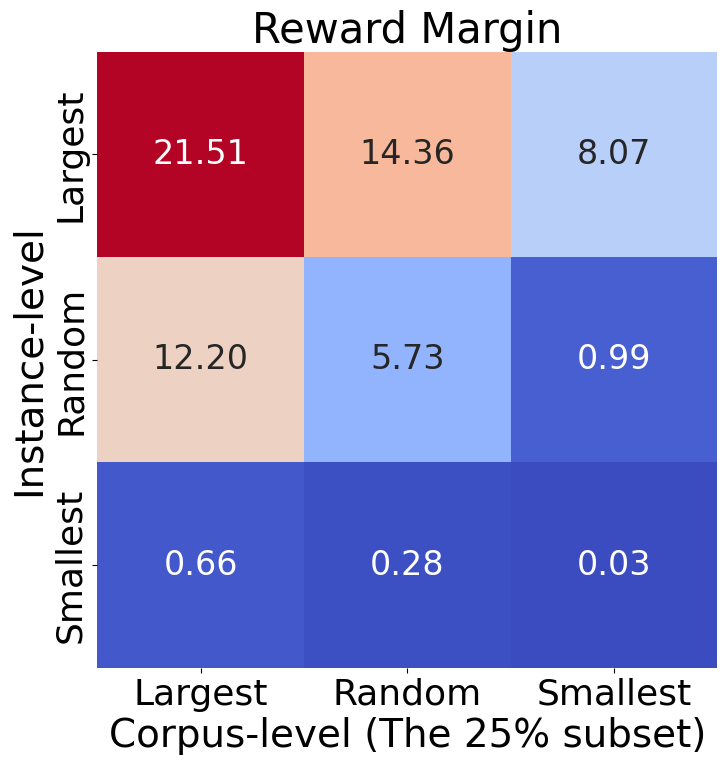}
    \caption{} 
    \label{subfig:exp:rq1:heatmap-margin}
\end{subfigure} \hspace{\fill} 
\begin{subfigure}[t]{0.26\textwidth}
    \includegraphics[width=\linewidth]{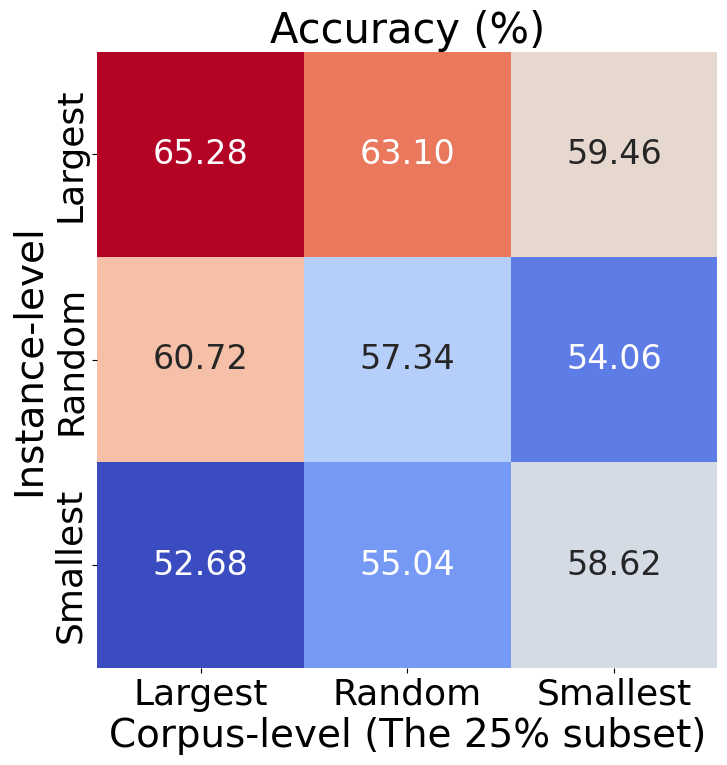}
    \caption{} 
    \label{subfig:exp:rq1:heatmap-acc}
\end{subfigure}
\hspace{\fill} 
\begin{subfigure}[t]{0.26\textwidth}
    \includegraphics[width=\linewidth]{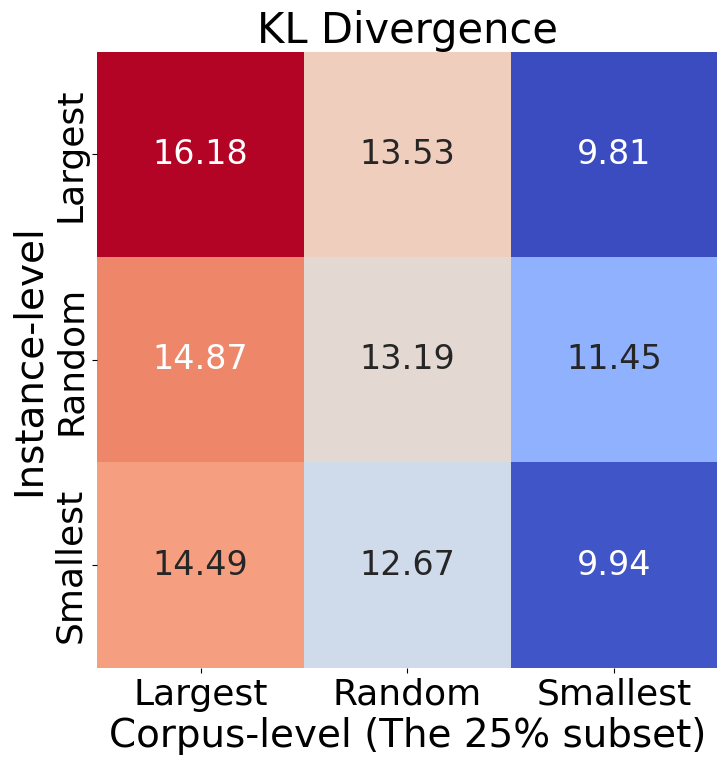}
    \caption{} 
    \label{subfig:exp:rq1:heatmap-kldiv}
\end{subfigure}
\caption{
        Some statistics about the selected subset using different strategies combined.
        The collected subsets are used to train $\pi_\theta^1$ under the single-iteration setting.
        }
\label{fig:exp:rq1:heatmap}
\end{figure*}

\subsection{Results}
\label{sec:rq1_results}

\paragraph{\textit{smallest} $>$ \textit{random} $>$ \textit{largest}}
We evaluate the single-iteration performances trained on different numbers of instances, with instance-level and corpus-level ranking.
The win rates against GPT-4 outputs on AlpacaEval, as evaluated by the gold RM and GPT-4, are shown in Figures~\ref{fig:exp:rq1:instance} and \ref{fig:exp:rq1:corpus}.
We first observe that annotating more data does not necessarily lead to more performance gain. This shows that it is necessary to strategically select data for annotating.
Selection with \textit{smallest} margins yields consistent improvements over the \textit{random} baseline, regardless of the instance- or corpus-level rankings.
In contrast, selection with \textit{largest} margins shows negative effects, which may be caused by overfitting to the confident instances.
Such results conform with our intuitions in $\S$\ref{sec:single_iter}.

\paragraph{Instance-level}
As shown in Figure~\ref{fig:exp:rq1:instance}, the three ranking schemes show consistent gaps across different numbers of samples, suggesting that it is not enough to just sample $N>2$ responses for a prompt but one also needs to pay attention to select the proper preference pair.
In our experiments, we show it is beneficial to select the preference pair with the \textit{smallest} margin among all $\frac{N(N-1)}{2}$ pairs.

\paragraph{Corpus-level}
In Figure~\ref{fig:exp:rq1:corpus}, the curve labeled as \textit{largest} significantly improves in terms of win rate between 15,000 and 20,000 samples.
The only difference between the two is adding the 5,000 samples with the \textit{smallest} margins to the training set.
In comparison, the curve labeled as \textit{smallest} shows a significant drop from 15,000 to 20,000 samples; and the only difference is including the \textit{largest}-margin 5,000 samples in the training set.
Taking both together, our experiments suggest including the \textit{smallest}-margin subset while discarding the \textit{largest}-margin subset during corpus-level data selection.

\paragraph{Length Normalization}
Table~\ref{tab:exp:rq1:avg_logp} shows the averaged win rates of the normalized models against the un-normalized ones across multiple runs.
We emphasize \textit{smallest} since we found it to be the best strategy for both levels of selection.
On the instance level, un-normalization shows clear improvement over the normalized counterpart; while on the corpus level, the superiority persists but is not significant.
Overall, our experiments suggest using the un-normalized reward formulation for both levels of granularity.



\begin{figure*}[th!]
\captionsetup[subfigure]{justification=Centering}
\centering
\begin{subfigure}[t]{0.42\textwidth}
    \includegraphics[width=\linewidth]{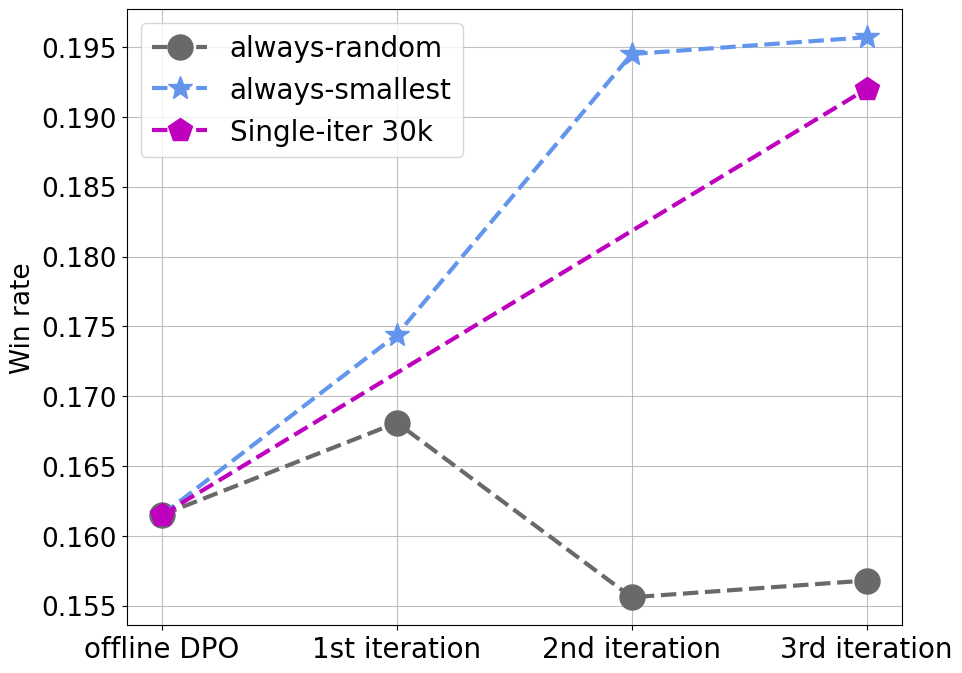}
    \caption{Evaluated the gold RM. } \label{subfig:exp:rq1:goldrm}
\end{subfigure} \hspace{\fill} 
\begin{subfigure}[t]{0.42\textwidth}
    \includegraphics[width=\linewidth]{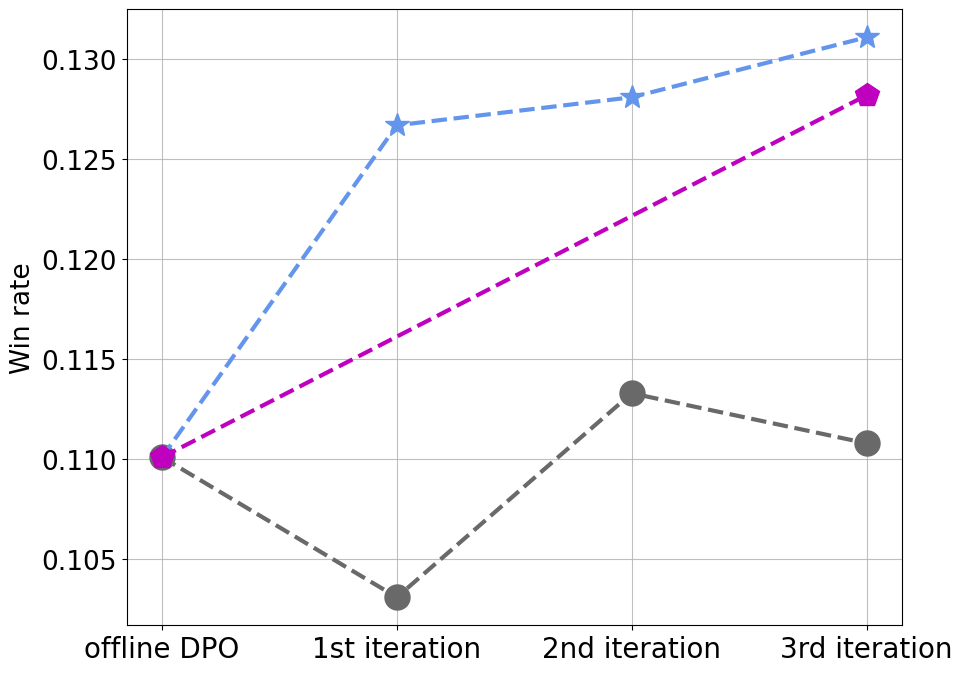}
    \caption{Evaluated by GPT-4. } \label{subfig:exp:rq2:gpt}
\end{subfigure}
\caption{
        Multi-iteration results on AlpacaEval-2.0 with \textit{always-random} and \textit{always-smallest} strategies, respectively, across three follow-up iterations, with 5k instances (originally 10k instructions) per iteration.
        A single-iter baseline, which is trained by using all the instructions with the \textit{always-smallest} strategy within a single round, is also included for comparison.
        }
\label{fig:exp:rq2}
\end{figure*}

\subsection{Analysis}
\label{subsec:rq1:analysis}

Figure~\ref{fig:exp:rq1:heatmap} shows some statistics on the selected subsets, including reward margin (as predicted by $\pi_\theta^0$ and $\pi_\mathrm{ref}$), ranking accuracy, and KL-divergence.
\paragraph{Uncertainty}
As shown in Figure~\ref{subfig:exp:rq1:heatmap-acc}, the accuracies generally follow a good calibration trend, i.e., ranking accuracy gradually increases as the reward margin increases.
For example, \textit{corpus-random instance-smallest} and \textit{corpus-smallest instance-random}, with ranking accuracies of $55.04\%$ and $54.06\%$, have been shown to yield better alignment performances than \textit{corpus-random instance-largest} and \textit{corpus-largest instance-random}, whose ranking accuracies are $63.10\%$ and $60.72\%$.
Therefore, our uncertainty assumption, that the model benefits more from training on the set with a higher degree of uncertainty, is partly supported.

\paragraph{Distribution Shift}
Since the reward function of DPO directly models the generative distribution of the LLMs, the reward margin measures the difference between the generative behaviors of the policy and the reference model, which may correspond to the distribution shift between the to-be-annotated instances and the already-trained ones.
We measure the degree of distribution shift using KL-divergences and sketch them on different groups of data points in Figure~\ref{subfig:exp:rq1:heatmap-kldiv}.
It is observed that there is a rough trend, with smaller reward margins more likely come smaller KL-divergences, i.e., smaller distribution shifts.
For example, the \textit{corpus-largest instance-largest} strategy, with a reward margin of $21.51$, yields a KL-divergence of $16.18$, which is much higher than the KL-divergences of other strategies that have smaller reward margins as well.

\section{Training for Multiple Iterations}
\label{sec:multi_iter}

Given our findings in the single-iteration case in $\S$\ref{sec:single_iter}, we would like to know how well the winning strategies generalize to multiple iterations (\textbf{Q4.1}).
Besides, recent work reported continuous improvements with multiple iterations for iterative DPO~\citep{wu2024selfplay,dong2024rlhf}.
They all adopted an evenly distributed strategy to allocate annotation budgets across iterations.
A natural question is, whether the model benefits from training on more instances in earlier or later iterations, rather than always training on the same amount (\textbf{Q4.2}).
The background experimental setup follows that of $\S$\ref{sec:setup:preliminary}.

\subsection{Empirical Workflow}
\label{sec:rq2:setup}

The multi-iteration workflow starts from $\pi_\theta^0$ and a set of instructions from UltraFeedback.
We divide the instructions into 3 sets for 3 rounds of iteration, where the $i$-th round uses $M_i$ instructions.
Specifically, for the $i$-th round of iteration, the following steps are implemented:
\paragraph{Step 1}
$N=8$ responses are sampled from $\pi_{\theta}^{i-1}$ for each instruction. 
$\pi_{\mathrm{ref}}$ and $\pi_{\theta}^{i-1}$ are used to predict the implicit reward, $\log \frac{\pi_{\theta}^{i-1}(y|x)}{\pi_{\mathrm{ref}} (y|x)}$.

\paragraph{Step 2}
We design experiments to answer the two questions, \textbf{Q4.1} and \textbf{Q4.2}.

\noindent\textbf{Q4.1: Always-smallest \textit{versus} \textbf{Always-random}}: 
    \begin{itemize}[noitemsep]
        \item \textbf{Always-smallest}: 
        We first select the response pair with the smallest reward margin among all $\frac{N(N-1)}{2}=28$ pairs per instruction.
        All the instructions and selected response pairs are collected to formulate a corpus, on which a corpus-level ranking is then applied to select the $50\% \times M_i = 5,000$ instances with the smallest reward margins.
        The selected instances are annotated by the gold RM and then used as the training set for the current iteration.
        We adopt the un-normalized reward formulation.
        
        \item \textbf{Always-random}:
        On the instance level, we simply select the first two responses to formulate a pair for each instruction.
        All the instructions and selected response pairs are collected to formulate a corpus, among which $50\% \times 10,000=5,000$ of the instances are randomly sampled.
        The sampled instances are fed into the gold RM for annotation and then used as the training set.
    \end{itemize}

\noindent\textbf{Q4.2: Increase \textit{versus} Constant \textit{versus} Decrease}:
    We adopt the \textit{always-smallest} strategy in this setup.
    After three rounds of iteration, each allocation strategy is trained on 30,000 instances.
    The numbers of instructions used for the three iterations for each case are as follows ($M_1 \rightarrow M_2 \rightarrow M_3$):
    \begin{itemize}[noitemsep]
        \item {Increase}: 5,000 $\rightarrow$ 10,000 $\rightarrow$ 15,000;
        \item Constant: 10,000 $\rightarrow$ 10,000 $\rightarrow$ 10,000;
        \item Decrease: 15,000 $\rightarrow$ 10,000 $\rightarrow$ 5,000.
    \end{itemize}

\paragraph{Step 3}
The collected subset is fed into $\pi_\theta^{i-1}$ and $\pi_{\mathrm{ref}}$ to obtain $\pi_\theta^i$, which then gets evaluated.

\subsection{Results}

\paragraph{Answer to Q4.1}
Figure~\ref{fig:exp:rq2} shows the win rates of \textit{always-smallest} and \textit{always-random}.
The \textit{always-random} baseline yields moderate improvements in the first one or two iterations but finally drops down upon being further optimized; while the \textit{always-smallest} strategy gives consistent and significant improvements across three iterations.
This suggests that the selection of response pairs for annotation plays a crucial role in facilitating continuous improvements for online iterative DPO.
Besides, the single-iter-30k baseline lags behind \textit{always-smallest}, indicating the effectiveness of corpus-level selection.


\vspace{-5pt}

\paragraph{Answer to Q4.2}
Table~\ref{tab:exp:rq2:allocation} shows the results with different allocation strategies.
Considering the results from both evaluators, \textit{decrease} is slightly better than \textit{constant} and much better than \textit{increase}.
This may result from the fact that the data quality in later iterations depends on the policy trained in earlier iterations, so it is better to allocate more data in the beginning to obtain a better policy.

\begin{table}[t]
    \centering
    \begin{adjustbox}{max width=0.9\columnwidth}
    \begin{tabular}{l|cc}
        \hline
        \diagbox{\textbf{Allocation}}{\textbf{Evaluator}}& {Gold RM} & {GPT-4} \\
         \hline
        Increase & 19.06 & 12.05 \\
        Constant & \textbf{19.57} & 13.11 \\
        Decrease & 19.41 & \textbf{13.49} \\
        
        \hline
    \end{tabular}
    \end{adjustbox}
    \caption{Results on different strategies (increase, constant, and decrease) to allocate annotation budgets across multiple iterations. We evaluate the win rates against GPT-4 outputs using the gold RM and GPT-4 as evaluators. The largest numbers are bolded. }
    \label{tab:exp:rq2:allocation}
\end{table}

\section{Related Work}
\label{sec:related_work}


\subsection{Iterative Preference Learning}
Online iterative preference learning refers to the framework in which response pairs are sampled from the policy models and are then annotated to become the training data to continuously improve the policy model.
Intuitively, it could mitigate the reward hacking issue or the distribution shift issue~\citep{Gao2022ScalingLF,rafailov2024scaling}
Online iterative preference learning has been verified to be effective for alignment methods with explicit rewards~\citep{bai2022training,llama2} and for direct preference learning methods~\citep{xu2024things,yuan2024selfrewarding,xiong2024iterative,rosset2024direct,wu2024selfplay,swamy2024minimaximalist,viethoangtranduong,ye2024online,guo2024direct,tajwar2024preference,calandriello2024human,chen2024bootstrappinglanguagemodelsdpo}.
Specifically, online direct preference learning was first presented in \citet{xu2024things}.
\citet{dong2024rlhf} shows a systematic training pipeline and releases a strong policy checkpoint based on \textsc{LLaMA}-3-8B-base.
Another line of work has investigated Nash equilibrium for LLM alignment, which is shown to natively support online iterative training by theory~\citep{wu2024selfplay,rosset2024direct,munos2024nash}.

\subsection{Active Learning for NLP}
Most active learning methods for NLP consider either \textit{informativeness}, such as prediction uncertainty~\citep{schroder-etal-2022-revisiting,margatina-etal-2021-active,zhang-etal-2022-allsh,jiang-etal-2020-camouflaged} and gradient~\citep{10.5555/2981562.2981724}, or \textit{representativeness}, such as representative of the unlabeled set~\citep{settles-craven-2008-analysis} and differences from already labeled instances~\citep{kim-etal-2006-mmr,zhao-etal-2020-active,erdmann-etal-2019-practical,gissin2019discriminative}.
In this work, we draw intuitions from both uncertainty (from the discriminative perspective) and representativeness (from the generative perspective).

\paragraph{Active Learning for LLM Alignment}
\citet{muldrew2024active} presented an active learning approach to make better use of a limited preference labeling budget.
Their methods are based on the assumption that the learning process is initialized from the base LLM and the annotated dataset is extremely small, thus the instances with large reward margins can provide greater gradients and alter the model's weights more significantly. 

\section{Conclusion}

In this work, we investigated strategies to make better use of limited annotation budgets for iterative preference learning.
Through extensive experiments, we found that it is better to select the response pairs with smaller predicted reward margins and to allocate more annotation budgets in earlier iterations.
We hope our findings could benefit the community to obtain better models with limited resources.

\section*{Limitations}
We observe several limitations regarding this work:
\begin{itemize}
    \item Our experiments are conducted under a synthetic-oracle setting.
    Though this setting has been widely adopted as a proxy of human oracle annotations~\citep{Gao2022ScalingLF,rafailov2024scaling} and we have included an external evaluator (GPT-4) to avoid the potential reward hacking issue, it is still possible that it introduces some unknown biases to the empirical findings.
    \item We did not consider the levels of annotating difficulty when measuring annotation cost.
    Throughout this work, we consider annotation cost to linearly correlate with the number of to-be-annotated response pairs.
    However, in practice, different pairs come with different levels of difficulty for annotators. 
\end{itemize}

\bibliography{acl_latex}

\clearpage

\appendix

\begin{table*}[ht]
    \centering
    \begin{tabular}{ccccc}
         \hline
         \textbf{Setting} & $\beta$ & Learning Rate & Batch Size & \# Epoch  \\
         \hline
         SFT & NA & 2e-5 & 128 & 1.0 \\
         DPO & 0.1 & 5e-7 & 128 & 1.0 \\
         \hline
    \end{tabular}
    \caption{Training Hyper-parameters. }
    \label{tab:appendix:train_hyper}
\end{table*}

\section{Implementation Details}
\label{sec:appendix:implmentation}

\paragraph{Training Setup}
Throughout this paper, we adopt the set of hyper-parameters shown in Table~\ref{tab:appendix:train_hyper}.
We use 8$\times$ A100-40G GPUs for all the training, with BF16 enabled.

\paragraph{Sampling Setup}
For sampling the responses for online training, we adopt a temperature of $1.0$ and top-$k$ sampling with $k=50$.
Both the max-prompt-length and the max-generate-tokens are set to $512$.
We enable BF16 during sampling.

\paragraph{Evaluation Setup}
We use greedy decoding to generate the responses for AlpacaEval-2.0.
Both the max-prompt-length and the max-generate-tokens are set to $512$.
We enable BF16 during generation.

\end{document}